# Prospective Algorithms for Quantum Evolutionary Computation

## Donald A. Sofge

Natural Computation Group
Navy Center for Applied Research in Artificial Intelligence
Naval Research Laboratory
Washington, DC, USA
donald.sofge@nrl.navy.mil

**Abstract**

This effort examines the intersection of the emerging field of quantum computing and the more established field of evolutionary computation. The goal is to understand what benefits quantum computing might offer to computational intelligence and how computational intelligence paradigms might be implemented as quantum programs to be run on a future quantum computer. We critically examine proposed algorithms and methods for implementing computational intelligence paradigms, primarily focused on heuristic optimization methods including and related to evolutionary computation, with particular regard for their potential for eventual implementation on quantum computing hardware.

## Introduction

A quantum computer is a device that processes information through use of quantum mechanical phenomena in the form of qubits, rather than standard bits as a classical computer does. The power of the quantum computer comes from its ability to utilize the superposition of states of the qubits. The potential of quantum computing (QC) over classical computing has received increasing attention due to the implications of Moore's Law on the design of classical computing circuits. Moore's Law (Moore, 1965) indicates that features sizes for integrated circuit components double about every 24 months. Current technology (Intel, 2008) provides high-volume production of 45nm processors in silicon, with 32nm silicon production capability targeted for 2009. However, nature places fundamental restrictions on how small certain devices may become before quantum mechanical effects begin to dominate. This will happen in the not-too-distant future.[1] By some estimates we will hit these limits by the year 2020, at which point transistor feature sizes will drop below 25nm, and putting greater numbers of transistors on a silicon wafer will require using larger and larger wafers, thereby also increasing power requirements and communication delays. Quantum computers offer a potential way forward through use of quantum mechanical states utilizing the principles of quantum superposition, interference, and entanglement. Quantum computers allow computation in a highly parallel manner through use of qubit representations and quantum gates where multiple solutions are considered simultaneously as superpositions of states, and interference is used to produce the desired solution.

Evolutionary computation (EC) represents a class of heuristic optimization techniques inspired by biological evolution that may be used to solve challenging practical problems in engineering, the sciences, and elsewhere. EC paradigms including genetic algorithms (GA), evolutionary strategies (ES), evolutionary programming (EP), genetic programming (GP), particle swarm optimization (PSO), cultural algorithms (CA), and estimation of distribution algorithms (EDAs) are population based search techniques generally run on digital computers, and may be characterized as searching a solution space for the fittest member of the population.

EC techniques are loosely based upon the Darwinian principle of "survival of the fittest". A solution space is modeled as a population of possible solutions that compete for survival from one generation to the next. Individual members of the population are represented according to their genomes (genotypic representation) and typically consist of strings of binary or real-valued parameters, although other representations such trees, linked lists, S-expressions (e.g. Lisp), may be useful as well, depending upon the target application.

The most common model of an EC method is the generational genetic algorithm. Various operators such as mutation, crossover, cloning, and selection strategies are applied to the individuals in the population, producing offspring for the following generation. The operators are designed to increase the fitness of the population in some measurable sense from one generation to the next such that

---

This work was supported by the Naval Research Laboratory under Work Order N0001406WX30002.

[1] A recent change in direction for semiconductor fabrication known as High-k focuses on use of alternate materials to silicon (such as hafnium-based compounds) to minimize quantum effects (Bohr *et al.*, 2007). This was called the first major redesign of the transistor in 40 years.

over a number of evolutionary cycles (generational runs) an individual or multiple individuals of superior fitness will be produced. These highly fit individuals represent more optimal solutions in the solution space, and are often the end product of using the EC technique to solve a challenging optimization problem.

Evolutionary computation has been successfully applied to a wide variety of problems including evolved behavior-based systems for robots, schedule optimization problems (including the traveling salesperson problem (TSP)), circuit design (including quantum circuit design, as will be discussed later), and many control optimization tasks (to name just a few). Heuristic optimization techniques are often most appropriately applied when: (a) the space being optimized defies closed-form mathematical description such that more direct optimization techniques could be used, (b) constraints or nonlinearities in the solution space complicate use of more direct optimization methods, (c) the size of the state space to be explored is exponentially large, precluding use of exhaustive search, and (d) exploration costs may high. A combination of any one or more of these factors may recommend use of heuristic EC techniques over more direct optimization methods such as line search, linear programming, Newton's method, or integer programming.

## Combining Quantum and Evolutionary Computation

The potential advantages of parallelism offered by quantum computing through superposition of basis states in qubit registers, and simultaneous evaluation of all possible represented states, suggests one possible way of combining the benefits of both quantum computing and evolutionary computation to achieve computational intelligence (e.g. better heuristic optimization). However, other notions have thus far resulted in more publications, and arguably greater success, in the EC and QC literature.

The first serious attempt to link quantum computing and evolutionary computation was the Quantum-inspired Genetic Algorithm (Narayanan & Moore, 1996). This work and several others to follow (described in greater detail below) focused on the use of quantum logic to inspire the creation of new algorithms for evolutionary computation to be run on classical computers. A substantial body of literature is now emerging called Quantum Interaction (Bruza et al., 2007) in which quantum models are used to explore new relationships and paradigms for information processing, including forms of computational intelligence (Laskey, 2007). Quantum-inspired evolutionary algorithms could be considered part of this branch.

The second serious undertaking combining quantum and evolutionary computation was perhaps inspired by the use of genetic programming to obtain novel circuit designs (Koza, 1997). Lee Spector (1998, 2004) explored the use of genetic programming to evolve new quantum algorithms, arguing that since quantum algorithms are difficult to create due to programmers' lack of intuitive understanding of quantum programming, then why not use GP to evolve new quantum algorithms. Spector was successful in evolving several new quantum algorithms in this fashion, though none as potentially useful or with such dramatic results as Grover's (Grover, 1996) or Shor's (Shor, 1994) algorithms (Nielsen & Chung, 2000).

The third combining of quantum and evolutionary computation, and the primary focus of this effort, is the one first mentioned in the section, leveraging the advantages of both quantum computing and evolutionary computation through EC algorithms to run (once suitable quantum hardware is available) on a quantum computer. We will focus on efforts reported in the literature to accomplish this, discuss their limitations, and then propose specific areas to focus research efforts to achieve this goal.

## Quantum-inspired Evolutionary Computation

### Quantum-Inspired Genetic Algorithm (QIGA)

In 1996 Narayanan & Moore published their work on the Quantum-inspired Genetic Algorithm (Narayanan & Moore, 1996) explicitly based upon the many-universes interpretation of quantum mechanics first proposed by Hugh Everett in 1957 (Everett 1957) and later espoused by David Deutsch (Deutsch 1997). In the quantum-inspired genetic algorithm (QIGA) each universe would contain its own population of chromosomes. This technique was discussed in terms of solving the traveling salesperson problem (TSP), and as such a necessary constraint was that no letter (representing a city to be visited) would be repeated in the string. The individual chromosomes are used to form a 2D matrix, with one chromosome per row representing the route and and with letters (representing the corresponding cities) in columns. An interference crossover operator is introduced that would form new individuals by selecting a new gene from the genome by following the diagonals (e.g. gene 1 is taken from the first element of universe 1, gene 2 from the second element in universe 2). In order to resolve possible replications in the string resulting from the interference crossover operator, if a letter (gene) is encountered which already occurs in the string, it is skipped and the pointer moves to that gene's immediate right (or wraps around if it is at the end of the string).

While this is an interesting model for evolutionary computation, there is little evidence presented why this would be better than other crossover or mutation techniques run on classical computers. One could surmise that the disruption factor would be extremely high, making this algorithm similar to random search over valid string formations. Other claimed advantages over traditional EC techniques would require further validation over a larger test suite than the one problem instance provided.

More significantly, the algorithm doesn't explain how the various representation and operators presented might be implemented using quantum logic gates, entanglement, how the interference between universes would manifest, and other quantum programming details. However, this would not be expected or necessary so long as there is no intention to run it on a quantum computer.

*Genetic Quantum Algorithm (GQA)*

In 2000 Han & Kim proposed the Genetic Quantum Algorithm (GQA) as a serious evolutionary algorithm to be run on a quantum computer. It uses the qubit as the basic unit of representation, and qubit registers are used to represent a linear superposition of all states that may occur in the chromosome. It suggests advantages over classical computation through parallelism (all individuals are processed simultaneously in parallel), and recognizes the need to use quantum gates for implementing the evaluation function (rotation gates are proposed). This algorithm is discussed in the context of solving the 0-1 knapsack problem, a known NP-Complete problem. However, the algorithm does not use mutation or crossover, and in fact suggests that use of such operators may decrease performance of the search.

```
procedure GQA
begin
     t ← 0
     initialize Q(t)
     make P(t) by observing Q(t) states
     evaluate P(t)
     store the best solution b among P(t)
     while (not termination-condition) do
     begin
          t ← t + 1
          make P(t) by observing Q(t - 1) states
          evaluate P(t)
          update Q(t) using quantum gates U(t)
          store the best solution among P(t)
     end
end
```

Figure 2. Pseudocode for QGA

Figure 2 shows the pseudocode for the algorithm. The population is stored in the qubit register Q, and it's state at time t is given by Q(t). Measurements on Q(t) are made and stored in the register P, with the measured states given by P(t).

The key difficulty with implementing this algorithm on a quantum computer is that it is generally not possible to measure all of the quantum states. Once a measurement is performed, the wave function collapses, and only the single measurement (which may be noisy and stochastic) is available.

The statement *make P(t) by observing Q(t-1)* collapses the state of Q into a single measurement, if P(t) is considered to be in a decoherent (or collapsed) state. If P(t) is considered at the time to still be in a superposition state, then make P(t) in fact represents a copying of Q(t) which violates the no-cloning theorem (Nielsen & Chung, 2000). Either way, the algorithm is infeasible as a quantum computer algorithm.

*Quantum Genetic Algorithm (QGA)*

In 2001 Bart Rylander, Terry Soule, James Foster, and Jim Alves-Foss proposed Quantum Evolutionary Programming and the Quantum Genetic Algorithm (QGA) (Rylander et al., 2001). This work built specifically upon notions of the quantum Turing machine, superposition, and entanglement. Again, this technique sought to leverage the parallelism of quantum computing for use in evolutionary search. Qubit registers were used to represent individuals in the population.

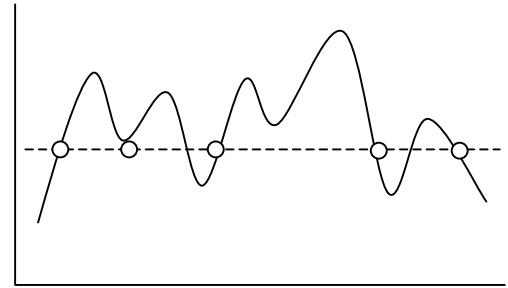

Figure 3. Fitness function for several classical individuals

One interesting feature shown in Figure 3 (recreated from Rylander's paper) is that several (but not all) classical members of a population are represented by a single quantum individual. Another interesting feature of this design was the use of two entangled quantum registers per quantum individual, the first an *Individual register*, and the second a *fitness register* as shown in Figure 4.

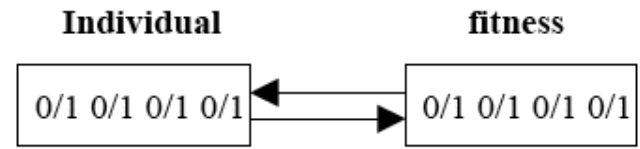

Figure 4. Two 4-qubit quantum registers are entangled

Rylander et al. recognize that the genotypic state of the individuals (and population) is not the same as the phenotypic (expressed fitness) state, and that there is a need to maintain entanglement between the two registers. The algorithm leaves out the crucial evaluation step (presumably performed by a quantum circuit composed of quantum gates) that would fall between the *Individual* and *fitness registers*, and most likely represent the most complex part of the algorithm.

Rylander et al. recognized several problems with their algorithm as a serious contender for an evolutionary algorithm to be run on a quantum computer. First, they recognized that observation (measurement) would destroy the superposition state (but no remedy was provided). Second, they recognized the limitation that the no-cloning theorem would put on any cloning or mutation operator (again, no remedy provided). Brushing these caveats aside, they argued that the larger effective population size due to superposition would lead to more efficient algorithms, and result in better building blocks. This last point seems particularly unsupported since without a sampling based method (e.g. all states are exhaustively searched) there is no need for building blocks. It's also hard to argue that building blocks are "better" due to use of a larger effective population size, since one could always just increase the population size, and the goal of exploiting building blocks is to restrict the size of the space to be searched.

*Quantum Evolutionary Algorithm (QEA)*

In 2004 Shuyuan Yang, Min Wang, and Licheng Jiao proposed the Quantum Evolutionary Algorithm (QEA) (Yang et al. 2004). This algorithm appears to be a synthesis of the QGA algorithm by Han & Kim and particle swarm optimization (Kennedy and Eberhart, 2001). QEA adopts the quantum crossover operator proposed by Narayanan & Moore, creating individuals for the new population by pulling elements from the diagonals in the previous population. In addition, they propose use of a guide chromosome whose update equations are given

$$Q_{guide}(t) = \alpha \times p_{currentbest}(t) + (1-a) \times (1 - p_{currentbest}(t)) \quad (9)$$
$$Q(t+1) = Q_{guide}(t) + b \times normrnd(0,1) \quad (10)$$

Figure 5. Update equations for Guide chromosome

These equations are very similar to the update equations used for calculating the update for particle swarm optimization. The resulting dynamics of this mutation operator should bear close resemblance to those of particle swarms.

*Quantum-Inspired Evolutionary Algorithm (QIEA)*

In 2004 Andre Abs da Cruz, Carlos Barbosa, Marco Pacheco, and Marley Vellasco proposed the quantum-inspired evolutionary algorithm (QIEA) to be run on a classical computer (Abs da Cruz 2004). This paper presents a quantum inspired evolutionary algorithm (actually more akin to an estimation of distribution algorithm (EDA) (Pelikan et al., 2006)) such that each variable represents a probability distribution function modeled by rectangular pulses. The center of each pulse represents the mean of each variable, and the pulse height is the inverse of the domain length/N, where N is the number of pulses used to encode the variable. The "quantum-inspiration" comes from the constraint that the sum of the areas under the N pulses must equal 1 (thus, the pulses represent a superposition). Probability distributions are altered each generation by updating the means (centers) and widths of each pulse by

a) randomly selecting m=n/N individuals (roulette method)

b) calculating the mean value of the m selected individuals and assigning it to a pulse, and
c) calculating the contraction of a pulse by

$$\sigma = (u-l)^{(1-\frac{t}{T})^{\lambda}} - 1$$

d) This technique bears considerable similarity to EDAs as mentioned previously. The technique is applied to optimization of the F6 function defined

$$F6(x,y) = 0.5 - \frac{(\sin\sqrt{x^2+y^2})^2 - 0.5}{1.0 + 0.001(x^2+y^2)^2}$$

While this technique is interesting, it proffers little toward the goal of finding an evolutionary algorithm to run on a quantum computer (indeed, that was not its purpose). The connection with quantum computing is minimal bordering on non-existent.

*Quantum Swarm Evolutionary Algorithm (QSE)*

In 2006 Yan Wang, Xiao-Yue Feng, Yan-Xin Huang, Dong-Bing Pu, Wen-Gang Zhou, Yan-Chun Liang, Chun-Guang Zhou introduced the quantum swarm evolutionary (QSE) algorithm based upon Han's GQA and particle swarm optimization as described by Kennedy and Eberhart (Kennedy and Eberhart, 2001). The particle swarm update equations are given by

$$v(t+1) = v(t) + c_1 * \text{rand}() * (p\text{best}(t) - \text{present}(t))$$
$$+ c_2 * \text{rand}() * (g\text{best}(t) - \text{present}(t)),$$
$$\text{present}(t+1) = \text{present}(t) + v(t+1).$$

Figure 6. QSE update equations for particles

where each particle maintains a state vector including spatial position and velocity. The present position of each is given by *present*, and the previous best position (based upon fitness) is given by *pbest* and the global best is given by *gbest*. The velocity term $v(t+1)$ is updated based upon several factors including the previous velocity $v(t)$, *pbest*, *gbest*, *present*, and two pre-selected constants $c_1$ and $c_2$. This algorithm is applied to the 0-1 knapsack problem and to the traveling salesman problem (TSP) with 14 cities. TSP and 0-1 knapsack are both known to be NP-Complete problems.

*Discussion*

The quantum-inspired techniques described in this section share a number of shortcomings for implementation on actual quantum hardware. Some of these shortcomings have been recognized and pointed out by the various authors, while others have not. A key question virtually ignored in all of these works is how the evaluation of fitness can be performed on a quantum computer. Performing the fitness evaluations for a large population of individuals is often the computational bottleneck facing use of EC techniques, and it is the area where one might hope that the massive parallelism of quantum computing might offer some benefits. However, most of the techniques discussed thus far do not address this evaluation piece. Moreover, if we are to leverage the benefits of quantum computing for evolutionary computation, then we need to find a way to perform measurements of the population of individuals without collapsing the superposition of fitness states. Furthermore, how can entanglement correlations between the individuals and their fitness values (with the evaluation performed in between) be maintained? A key issue either not addressed or not resolved for all of these proposed techniques is how to implement mutation and crossover operators in quantum algorithms. Some of the papers suggest that since they are performing exhaustive search of the state space, it no longer makes sense to use such operators. This is logical, but it presents the question of why one should bother with a heuristic sample-based search strategy such as offered by EC techniques at all when in fact an exact method is available (that is, exhaustive search). If exhaustive search is available at the same computational cost as heuristic sample-based search, then there is no reason not to do the exhaustive search. If there does still exist some rationale for doing sample-based heuristic search, and one chooses an EC-like approach using operators such as crossover, mutation, and cloning, then how do you accommodate the no-cloning theorem? And how might such operators be built?

## "True" Quantum Algorithms for Optimization and Search

Given the difficulties with many of the quantum genetic algorithms derived from or inspired by classical techniques, one might try the approach of starting with quantum algorithms known or widely believed to be feasible or "true" quantum algorithms in the sense that if we had a large enough quantum computer, and given our current state of knowledge about quantum programming, it would be possible to implement such an algorithm on the quantum computer and run it. A reasonable place to start is with Grover's Algorithm, especially since it is a search algorithm.

*Quantum Genetic Algorithm based on Grover's Algorithm*
In 2006 Mihai Udrescu, Lucian Prodan, and Mircea Vladutiu proposed implementing a feasible quantum genetic algorithm based upon known quantum building blocks, especially Grover's Algorithm (Udrescu et al., 2006). Specifically, they begin by acknowledging that all previous designs for genetic algorithms running on quantum computers are infeasible (for reasons discussed above). They then focus attention on designing a quantum oracle to perform the evaluation function (mapping individuals to fitness scores), and then coupling it with a quantum maximum finding algorithm (Ahuja and Kapoor, 1999) (a variant of Grover's Algorithm (Grover 1996)) to reduce the process to a Grover's search. They point out that if the qubit representation can represent all possible states in the population, then there is no need for genetic operators such as crossover and mutation. The result is what they call the Reduced Quantum Genetic Algorithm (RQGA) (which is not truly a genetic algorithm at all, since it doesn't use genetic operators or perform sample-based optimization).

RQGA is constructed by beginning with Rylander's Quantum Genetic Algorithm QGA (Rylander et al., 2001) as shown in Figure 7. QGA first puts the members of the population into superposition in quantum (qubit) registers, then measures the fitness states (presumably without collapsing the superposition). Once the fitness measurements are available, then selection is performed according to measured fitness values, crossover and mutation are employed to produce new offspring (not explained how), new fitness values are computed. This process repeats until the desired termination condition is reached. (It is not difficult to find multiple reasons why this algorithm is infeasible).

QGA is in fact integrated with a quantum algorithm for finding the maximum (Ahuja and Kapoor, 1999) shown in Figure 8, to produce RQGA shown in Figure 9. Note that Grover's Algorithm is a subroutine used by the Maximum Finding algorithm.

**Genetic Algorithm Running on a Quantum Computer (QGA) with proper formalism**

1. For $i := 1$ to $m$ set the individual-fitness pair registers as $|\psi\rangle_i^1 = \frac{1}{\sqrt{n}} \sum_{u=0}^{n-1} |u\rangle_i^{ind} \otimes |0\rangle_i^{fit}$ (a superposition of $n$ individuals with $0 \leq n \leq 2^N$).

2. Compute the fitness values corresponding to the individual superposition, by applying a unitary transformation $U_{f_{fit}}$ (corresponding to pseudo-classical Boolean operator $f_{fit}$: $\{0,1\}^N \rightarrow \{0,1\}^M$). For $i := 1$ to $m$ do $|\psi\rangle_i^2 = U_{f_{fit}} |\psi\rangle_i^1 = \frac{1}{\sqrt{n}} \sum_{u=0}^{n-1} |u\rangle_i^{ind} \otimes |f_{fit}(u)\rangle_i^{fit}$.

3. For $i := 1$ to $m$ measure the fitness registers, obtaining the post-measurement states (we suppose that $|y\rangle_i$ is obtained by measurement): $|\psi\rangle_i^3 = \frac{1}{\sqrt{k_i}} \sum_{v \in \{0,1,\ldots,n-1\}} |v\rangle_i^{ind} \otimes |y\rangle_i^{fit}$ with $k_i$ values in $\{0,\ldots,n-1\}$ to satisfy $f_{fit}(v) = y$.

4. Repeat
   a. Selection according to the $m$ measured fitness values $|y\rangle_i$.
   b. Crossover and mutation are employed in order to prepare a new population (setting the $m$ individual registers $|u\rangle_i^{ind}$).
   c. For the new population, the corresponding fitness values will be computed and then stored in the fitness registers($|f_{fit}(u)\rangle_i^{fit}$).
   d. Measure all fitness registers

Until the condition for termination is satisfied.

Figure 7. QGA algorithm by Rylander et al.

**Quantum Algorithm for finding the maximum from an unsorted table of $m$ elements**

1. Initialize $k := random\ number;\ 0 \leq k \leq m - 1$ as the starting index of this search;
2. Repeat $\mathcal{O}\left(\sqrt{m}\right)$ times
   a. Set two quantum registers as $|\psi\rangle = \frac{1}{\sqrt{m}} \sum_{i=0}^{m-1} |i\rangle|k\rangle$; the first register is a superposition of all indices;
   b. Use Grover's algorithm for finding marked states from the first register (i.e. those which make $O_k(i) = 1$);
   c. Measure the first register. The outcome will be one of the basis states which are indices for values $> P[k]$. Let the measurement result be $x$. Make $k := x$;
3. Return $k$ as result. It is the index of the maximum.

Figure 8. Algorithm for Finding the Maximum (Ahuja and Kapoor, 1999)

**Reduced Quantum Genetic Algorithm**

1. For $i := 0$ to $m - 1$ set the pair registers as
   $|\psi\rangle_i^1 = \frac{1}{\sqrt{2^N}} \sum_{u=0}^{2^N-1} |u\rangle_i^{ind} \otimes |0\rangle_i^{fit}$;
2. For $i := 0$ to $m - 1$ compute the unitary operation corresponding to fitness computation
   $|\psi\rangle_i^2 = U_{f_{fit}} |\psi\rangle_i^1 = \frac{1}{\sqrt{2^N}} \sum_{u=0}^{2^N-1} |u\rangle_i^{ind} \otimes |f_{fit}(u)\rangle_i^{fit}$;
3. $max := random\ integer$, so that $2^{M+1} \leq max \leq 2^{M+2} - 1$;
4. For $i := 0$ to $m - 1$ loop
   (a) Apply the oracle $\tilde{O}_{max}\left(f_{fit}(u)\right)$. Therefore, if $|f_{fit}(u)\rangle_i^{fit} > max$ then the corresponding $|f_{fit}(u)\rangle_i^{fit}$ basis states are marked;
   (b) Use Grover iterations for finding marked states in the fitness register after applying the oracle. We find one of the marked basis states $|p\rangle = |f_{fit}(u)\rangle_i^{fit}$, with $f_{fit}(u)\ max \geq 0$;
   (c) $max := p$;
5. Having the highest fitness value in the $|\bullet\rangle_{m-1}^{fit}$ register, we measure the $|\bullet\rangle_{m-1}^{ind}$ register in order to obtain the corresponding individual (or one of the corresponding individuals, if there is more than one solution).

Figure 9. Reduced Quantum Genetic Algorithm (RQGA)

Note that RDQA assumes the existence of a quantum oracle (referenced in step 4(a)) to perform the fitness evaluation. This is a reasonable assumption given that no prior information is given about *what* the evaluation function actually *is*. The fitness states (and individuals) are kept in superposition, but a marking mechanism is used as with Grover's Algorithm. Step 4(a) applies the oracle and marks all of the basis states grater than *max*. Step 4(b) uses Grover iterations for finding marked states in the fitness register. One of the marked basis states is selected and copied to *p* (a measurement of f). {*No explanation is given of how this wouldn't cause collapse of the wave function*}. The loop repeats m times (m is the number of elements in the table), and the item(s) having the largest fitness value(s) in the *fitness register* are measured in the *individual register* to identify which individual(s) have the greatest fitness.

Udrescu et al. conclude that with quantum computation, genetic (heuristic sample-based) search becomes unnecessary. Furthermore, they point out that the computational complexity of RQGA is the same as Grover's Algorithm, $O(\sqrt{N})$. They suggest that if the quantum algorithm can be devised to exploit the structure of the problem space, then the search may be made faster.

Machine learning techniques (successful ones, at least) obtain an advantage over random search by exploiting the existence of structure in the problem domain, whether it is known *a priori* or discovered by the learning algorithm. It is well-known that finding a needle-in-a-haystack, that is, searching a problem domain with no structure other than random ordering, no strategy is more efficient than random search. Similarly, Grover's Algorithm has been shown to be optimal for searching unordered lists. If the list is ordered, or possesses some other exploitable structure (a more appropriate problem class for EC algorithms), then one might hopefully do better than random search, or in this case Grover's Algorithm.

*Other Special-Case Quantum Search*

Several quantum algorithms have been described in the literature that build upon prior quantum search techniques (usually Grover's Algorithm) and have superior performance to classical techniques if certain conditions are met.

In 1998 Tad Hogg (Hogg, 1998) proposed a quantum algorithm to solve the Boolean satisfiability problem. This is the problem of determining if the variables of a given Boolean formula can be assigned in such a way as to make the formula evaluate to TRUE. His technique (known as Hogg's Algorithm) implements Walsh transforms with unitary matrices representing transform, based on Hamming distances, and is capable of solving the 1-SAT or maximally constrained k-SAT (where k is the number of variables in the expression). Further, he showed that the best classical algorithms that exploit problem structure grow linearly with problem size, whereas both classical and quantum methods that ignore problem structure grow exponentially.

In contrast with Grover's Algorithm which is designed to search unordered lists, in 2007 Childs, Landahl, and Parrilo (Childs et al., 2007) examined quantum algorithms to solve the Ordered Search Problem (OSP), the problem of finding the first occurrence of a target item in an ordered list of N items, provided that the item is known to be somewhere in the list. They showed that the lower bound for classical algorithms (assuming that the list must be searched, instead of being indexed such as through a hash lookup) is $O(\log_2 N)$ {achieved by binary search}. They show that a quantum algorithm can improve upon this by a constant factor, and that the best known lower bound is $0.221*\log_2 N$. Further, they demonstrate a quantum OSP algorithm that is $0.433*\log_2 N$, the best showing to date.

In 2007 Farhi, Goldstone, and Gutmann (Farhi et al., 2007) devised a quantum algorithm for the binary NAND tree problem in the Hamiltonian oracle model. The Quantum Hamiltonian oracle model is special case of general quantum oracle model where elements of the tree form orthogonal subspaces. They showed that the computational complexity of the algorithm is $O(\sqrt{N})$, whereas the best classical algorithm is $N^{0.753}$. This result has subsequently been generalized to the standard quantum query model.

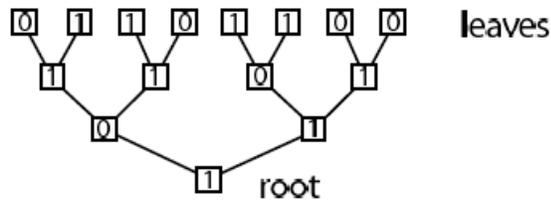

Figure 10. A Classical NAND Tree

## Summary and Conclusions

Evolutionary Computation and Quantum Computation both offer tremendous potential, but it isn't clear how to maximally take advantage of both simultaneously. Naïve approaches based upon known EC techniques will probably not work, until basic quantum search algorithms are better understood. It isn't currently known how to perform crossover operations such that entangled states will constructively interfere with one another.

Quantum oracles for searching specialized data structures need further exploration and development, and quantum programming techniques such as amplitude amplification need to be better understood to provide useful quantum subroutines for an increasing variety of data structures.

Often the most computationally demanding part of EC is not operations such as mutation and crossover, but in providing evaluations (e.g. by a quantum oracle) to generate fitness values. It is clear that for individuals in the population represented in superposed states, maintaining a correlation between genotypic (bitstring, e.g.) and phenotypic (expression/fitness) data will be critical to leveraging the potential advantages of QC.